%% file: main.tex
\documentclass[runningheads]{llncs}

\makeatletter
\def\input@path{{../}}
\makeatother

\usepackage{eccv}

\usepackage[T1]{fontenc}
\usepackage{lmodern}

\input{preamble}
\input{acronyms}
\input{paper_macros}

\usepackage{eccvabbrv}

\usepackage[accsupp]{axessibility}  %

\usepackage[breaklinks,colorlinks,citecolor=eccvblue]{hyperref}

\usepackage{orcidlink}

\begin{document}

\title{Diffusion-Based Environment-Aware \\ Trajectory Prediction} %

\author{Theodor Westny\inst{1}\orcidlink{0000-0001-9075-7477} \and
Björn Olofsson\inst{2,1}\orcidlink{0000-0003-1320-032X} \and
Erik Frisk\inst{1}\orcidlink{0000-0001-7349-1937}}

\authorrunning{T.~Westny \etal}

\institute{Department of Electrical Engineering, Linköping University \\
\email{\{theodor.westny, bjorn.olofsson, erik.frisk\}@liu.se}
\and
Department of Automatic Control, Lund University
\email{bjorn.olofsson@control.lth.se}}

\maketitle

\input{sec/0_abstract}    
\input{sec/1_introduction}
\input{sec/2_related_work}
\input{sec/3_model.tex}
\input{sec/4_results.tex}

\section{Conclusions}

In this paper, we have proposed a generative model for trajectory prediction based on the diffusion process.
The model is capable of capturing the complex interactions between traffic participants and the environment, accurately learning the multimodal nature of the data.
The effectiveness of the approach was assessed on large-scale datasets of real-world traffic scenarios, showing that our model outperforms several well-established methods in terms of prediction accuracy.
By the incorporation of differential constraints, we illustrate that our model is capable of generating a diverse set of realistic future trajectories, which is essential for the safe and efficient operation of autonomous vehicles.
Finally, Using a guidance signal based on node connectivity, we demonstrated that the model can be steered to generate predictions that are more or less influenced by the inter-agent interaction---an attribute we hypothesize to be beneficial in various real-world applications.

\section*{Acknowledgements}
This research was supported by the Strategic Research Area at Linköping-Lund
in Information Technology (ELLIIT) and the Wallenberg AI, Autonomous Systems and Software Program (WASP) funded by the Knut and Alice Wallenberg Foundation.
Computations were enabled by the Berzelius resource provided by the Knut and Alice Wallenberg Foundation at the National Supercomputer Centre.

\bibliographystyle{splncs04}
\bibliography{references.bib}

\appendix
\input{sec/A_experiments}
\newpage
\input{sec/B_diffusion}

\end{document}

%% file: preamble.tex
\usepackage{url}            %
\usepackage{xfrac}       %
\usepackage{booktabs}       %
\usepackage{microtype}      %
\usepackage{xcolor}         %
\usepackage{textcomp}

\usepackage{amsmath}
\usepackage{amssymb}
\usepackage{amsfonts}
\usepackage{bm}
\usepackage{mathtools}
\usepackage{cancel}

\usepackage{enumitem}
\usepackage{pifont}

\usepackage{wrapfig}
\usepackage{float}
\usepackage{graphicx}
\usepackage{subcaption}

\usepackage[ruled,vlined,linesnumbered]{algorithm2e}
\usepackage{algcompatible}
\usepackage{enumitem}

\usepackage[acronym]{glossaries}

%% file: acronyms.tex
\newacronym{GNN}{GNN}{graph neural network}
\newacronym{CNN}{CNN}{convolutional neural network}
\newacronym{MDN}{MDN}{mixture density network}
\newacronym{CVAE}{CVAE}{conditional variational autoencoder}
\newacronym{GAN}{GAN}{generative adversarial network}

\newacronym{MHA}{MHA}{multi-head attention}
\newacronym{GGRU}{Graph-GRU}{graph-gated recurrent unit}
\newacronym{GRU}{GRU}{gated recurrent unit}

\newacronym{HD}{HD}{high-definition}

\newacronym{ADE}{ADE}{Average Displacement Error}
\newacronym{FDE}{FDE}{Final Displacement Error}
\newacronym{MR}{MR}{Miss Rate}

\newacronym{GCN}{GCN}{graph convolutional network}
\newacronym{GAT}{GAT}{graph attention network}

%% file: paper_macros.tex
\newcommand{\highd}{\emph{highD}}

\newcommand{\round}{\emph{rounD}}

\newcommand{\R}{\mathbb{R}}
\newcommand{\set}[1]{\left\{#1\right\}}

\newcommand{\normal}{\mathcal{N}} %
\newcommand{\uniform}{\mathcal{U}} %

\definecolor{MyGreen}{HTML}{10A23B}
\definecolor{MyRed}{HTML}{DD1111}

\newcommand{\mtp}{MTP-GO}
\newcommand{\tplusplus}{Trajectron++}
\newcommand{\gplusplus}{GRIP++}
\newcommand{\mmtf}{mmTransformer}

\newcommand{\graph}{\mathcal{G}}
\newcommand{\node}{\mathcal{V}}
\newcommand{\edge}{\mathcal{E}}

\newcommand{\agent}{\nu}
\newcommand{\neighbor}{\tau}
\newcommand{\neigh}[1]{\mathbb{N}(#1)}
\newcommand{\ineigh}[1]{\tilde{\mathbb{N}}(#1)}

\newcommand{\feature}[2][\agent{}]{\bm{f}_{#2}^{#1}}

\newcommand{\inp}[1]{\bm{u}_{#1}}

\newcommand{\history}{\mathcal{H}}

\newcommand{\predhist}{h}
\newcommand{\predhrz}{N}

\newcommand{\lanegraph}{\mathcal{J}}

\newcommand{\hidden}{\bm{h}}

\newcommand{\encrep}[2][\agent]{\hidden_{#2}^{#1}}
\newcommand{\grurepx}[2][\agent]{\bm{\kappa}_{#2,i}^{#1}}
\newcommand{\grureph}[2][\agent]{\bm{\xi}_{#2,i}^{#1}}

\newcommand{\gnnf}[2][\agent{}]{\text{GNN}_{f}\left(\feature[#1]{#2}, \set{\feature[\tau]{#2}}_{\tau \neq{} #1} \right)}
\newcommand{\gnnh}[2][\agent{}]{\text{GNN}_{h}\left(\encrep[#1]{#2}, \set{\encrep[\tau]{#2}}_{\tau \neq #1} \right)}
\newcommand{\updatedrep}{{\hidden{}'}^{\agent{}}}

\newcommand{\ew}[2]{\bm{e}^{#1,#2}}

\newcommand{\gattw}[2][\agent{}]{\alpha_{\agent{},#2}}
\newcommand{\attn}{\bm{a}}
\newcommand{\weight}{\bm{W}}
\newcommand{\weightind}[1]{\weight_{#1}}

\newcommand{\transpose}{\text{T}} %

\newcommand{\al}[2][my_equation]{\begin{align}\label{eq:#1}#2\end{align}}

\newcommand{\lrelu}{\text{LeakyReLU}}

\newcommand{\zero}{\bm{0}}
\newcommand{\eye}{\bm{I}}

\newcommand{\model}{\mathcal{M}_\theta}

\newcommand{\alphat}[1]{\alpha_{#1}}
\newcommand{\alphabart}[1]{\bar{\alpha}_{#1}}

\newcommand{\schedule}{\gamma}

\newcommand{\betat}[1]{\beta_{#1}}

\newcommand{\joint}{p_\theta}
\newcommand{\posterior}{q}
\newcommand{\predlatent}[1]{\hat{\bm{x}}_{#1}}
\newcommand{\latent}[1]{\bm{x}_{#1}}
\newcommand{\noise}{\bm{\epsilon}}
\newcommand{\cond}{\bm{c}}

\newcommand{\dstep}{t}
\newcommand{\ddiff}{\mathrm{\Delta}}
\newcommand{\dsteps}{T}

\newcommand{\posteriormu}{\bm{\mu}_\dstep}
\newcommand{\posteriorvar}{\sigma_\dstep^2}

\newcommand{\varposteriormu}{\tilde{\bm{\mu}}_\dstep}
\newcommand{\varposteriorvar}{\tilde{\sigma}_\dstep^2}

\newcommand{\lookaheadconstant}{a}
\newcommand{\lookahead}{l}
\newcommand{\curvature}{c}

%% file: sec/0_abstract.tex
\begin{abstract}
The ability to predict the future trajectories of traffic participants is crucial for the safe and efficient operation of autonomous vehicles. In this paper, a diffusion-based generative model for multi-agent trajectory prediction is proposed.
The model is capable of capturing the complex interactions between traffic participants and the environment, accurately learning the multimodal nature of the data.
The effectiveness of the approach is assessed on large-scale datasets of real-world traffic scenarios, showing that our model outperforms several well-established methods in terms of prediction accuracy.
By the incorporation of differential motion constraints on the model output, we illustrate that our model is capable of generating a diverse set of realistic future trajectories.
Through the use of an interaction-aware guidance signal, we further demonstrate that the model can be adapted to predict the behavior of less cooperative agents, emphasizing its practical applicability under uncertain traffic conditions. 
\keywords{Trajectory Prediction $\bm{\cdot}$ Generative Modeling $\bm{\cdot}$ Autonomous Driving}
\end{abstract}

%% file: sec/1_introduction.tex
\section{Introduction}
An important property of self-driving vehicles is their ability to accurately forecast the behaviors (motion and intentions) of surrounding road users.
However, merely making accurate predictions is not sufficient in practical applications.
To enable safe and robust decision-making processes, it is crucial that these methods also effectively handle the multimodal uncertainty in the forecasted behaviors.
As a consequence, probabilistic modeling has become a cornerstone in devising robust trajectory prediction methods~\cite{huang2022survey}.
With the emerging field of generative modeling offering promising new directions for research and development, we seek to explore the potential of conditional diffusion models for trajectory prediction.

Diffusion models~\cite{sohl2015deep,ho2020denoising,song2021denoising,karras2022elucidating}, %
a type of generative model, have seen a surge in popularity in recent years.
These models have made a significant impact in the field of image synthesis~\cite{ho2020denoising,song2021denoising,ramesh2022hierarchical,rombach2022high}; and more recently, been innovatively applied in other domains, including molecule generation~\cite{hoogeboom2022equivariant}, temporal data modeling~\cite{rasul2021autoregressive,alcaraz2022diffusion}, traffic scenario generation~\cite{pronovost2023scenario}, and more~\cite{yang2023diffsurvey}. 
However, their adaptation and use for trajectory prediction applications are still limited. %

An important aspect of trajectory prediction, specifically multi-agent methods, is the modeling of inter-agent interactions. 
Using \glspl{GNN} for these purposes has emerged as one fundamental technique~\cite{huang2022survey,rahmani2023survey}.
Their scalability and flexibility in handling geometric data provide a distinctive inductive bias, ideal for modeling the complex interactions that are inherent in many real-world traffic scenarios.
While the modeling of inter-agent interactions has become a cornerstone in behavior prediction methods~\cite{alahi2016social,deo2018convolutional,li2019grip,messaoud2020attention,liu2021multimodal,westny2023mtp,wang2023spatio}, incorporating map-based information and road-agent interactions have likewise attracted the interest of several researchers~\cite{zhao2019multi,liang2020learning,messaoud2021trajectory,salzmann2020trajectron,li2021spatio,deo2022multimodal,hu2022scenario,zhou2023query}.
Unifying these approaches as conditionals in a diffusion model is the focus of this work.

In motion forecasting tasks, a notable drawback of black-box (neural) models is their tendency to compute outputs that either defy physical feasibility or lack the characteristics of natural motion.
To mitigate these issues, there have been several proposals for incorporating differential motion constraints within trajectory prediction models~\cite{cui2020deep,salzmann2020trajectron,li2021spatio,yue2022human,wen2022social,westny2023mtp,westny2023eval}.
Still, their direct inclusion in the diffusion framework is not appropriate given the original denoising objective~\cite{ho2020denoising}.
However, by a simple modification of the model output and learning objective, we can directly enforce physical properties in the learnable model, enabling realistic and physically feasible trajectory predictions.
\begin{figure}[!t]
    \centering
    \includegraphics[width=\textwidth]{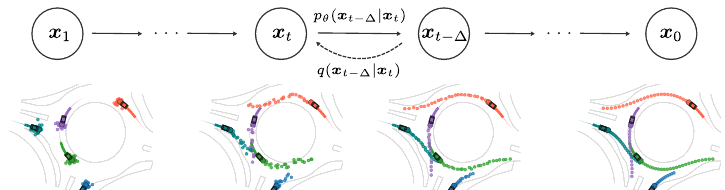}
    \caption{The directed graphical model considered in this work. The goal of diffusion models is to learn the process of transforming noise into samples that are representative of the true data distribution.
    In this work, the task of the proposed model is to generate realistic and physically feasible future trajectories for road users.
    }
    \label{fig:graphical-model}
\end{figure}
\subsection{Contributions}
The primary contributions of this paper are:
\begin{itemize}[label=\ding{228}]
	\item A diffusion-based multi-agent trajectory prediction model that conditions predictions on inter-agent interactions and map-based information.
    It generates physically feasible predictions by the use of differential motion constraints.
	\item A comprehensive evaluation and comparison of the proposed model on two large-scale real-world datasets, showcasing its effectiveness in handling multimodality and uncertainty in trajectory predictions.
	\item An investigation into guided sampling based on inter-agent connectivity and its effect on the prediction process.
\end{itemize}
Investigations were conducted using the \highd{}~\cite{highDdataset} and \round{}~\cite{rounDdataset} datasets.
Implementations will be made available online.

%% file: sec/2_related_work.tex
\section{Related Work}
\label{sec:related_work}
In this section, we review works that are close to our proposal.
For more general overviews of trajectory prediction methods, we refer to several extensive surveys on the topic, \eg, \cite{fang2022behavioral,huang2022survey}.

\subsection{Probabilistic Modeling} 
Considering the complex dynamics of traffic and the multiple possible decisions of individual road users, it is no coincidence that several researchers have employed approaches with probabilistic elements toward the solution of the behavior prediction problem.
One common method is to model future trajectories as a distribution over possible outcomes~\cite{hu2018probabilistic,deo2018convolutional,messaoud2020attention,messaoud2021trajectory,westny2023mtp,zhou2023query} using \glspl{MDN}~\cite{bishop1994mixture}.
Contrasting these works are those that employ generative approaches, such as \glspl{GAN}~\cite{goodfellow2014generative} to implicitly encode multimodality~\cite{gupta2018social,amirian2019social,sadeghian2019sophie,zhao2019multi} or using \glspl{CVAE}~\cite{sohn2015learning} to explicitly do so~\cite{salzmann2020trajectron,yuan2021agentformer,li2021spatio,xu2022socialvae}. %

Given the recent success of diffusion models in multiple domains~\cite{yang2023diffsurvey}, it is natural to consider their potential for trajectory prediction.
Although there are notable examples of using diffusion-based models in temporal data modeling~\cite{rasul2021autoregressive,alcaraz2022diffusion}, employing them specifically for motion generation has largely been focused on computer graphics applications~\cite{zhang2022motiondiffuse,kim2023flame,tevet2023human,rempe2023trace}.
While these works offer important insights, they are typically single-agent methods and therefore lack the interaction-aware mechanisms that have been the recipe for success in many recent (multi-agent) trajectory prediction approaches. %
Of particular interest are therefore diffusion models that in addition to modeling temporal mechanisms also incorporate spatial components, \eg, \glspl{GNN}, such as in~\cite{wen2023diffstg,liu2023pristi,li2023graph}.

Despite recent developments, the research on diffusion-based models tailored for trajectory prediction remains an open area of research, with only a few works exploring their potential~\cite{gu2022stochastic,chen2023equidiff,rempe2023trace}.
In~\cite{gu2022stochastic}, the authors proposed a Transformer-based diffusion model for pedestrian trajectory prediction.
Showing promising results across several pedestrian datasets, they specifically highlight the ability of the model to capture multimodal behavior.
Despite these promising results, the authors note the time cost of the reversal process, which is a known issue with diffusion models~\cite{song2021denoising,karras2022elucidating} (this was later addressed in the public release of the code).
In~\cite{chen2023equidiff}, the authors proposed a diffusion-based model for the prediction of vehicle trajectories on highway driving datasets.
The model combines a transformer with recurrent mechanisms and a \gls{GNN} to capture temporal and spatial dependencies. 
The authors show that the model outperforms several baselines on short-term prediction tasks, but struggles to accurately predict longer trajectories.
In~\cite{rempe2023trace}, a diffusion-based model was proposed for the prediction of pedestrian trajectories for character animation.
In the work, the authors proposed a two-stage approach, where the first stage incorporates a diffusion-based model that predicts the future trajectory, subsequently used as a plan for a physics-based controller in the second stage.
The key contribution of the work is the added controllability which enables guidance at test time.

\subsection{Environment-Aware Modeling}
Since the seminal work in~\cite{alahi2016social}, several researchers have targeted the problem of developing interaction-aware models with various choices of architectures, including pooling techniques~\cite{alahi2016social,gupta2018social,deo2018convolutional}, \glspl{CNN}~\cite{deo2018convolutional,zhao2019multi,cui2019multimodal}, and attention mechanisms~\cite{amirian2019social,messaoud2020attention,park2020diverse,liu2021multimodal,huang2022multi}.
However, the natural representation of trajectories as temporal graphs, where nodes represent entities or key locations, and edges represent their relationships has enabled \glspl{GNN} to emerge as the most commonly used technique.
In \cite{diehl2019graph}, one of the first works to use \glspl{GNN} for motion prediction, the authors investigated selected architectures, presenting promising results in terms of interaction-aware modeling.
In response, there have been several proposals that include \glspl{GNN} within the prediction model~\cite{li2019grip,jeon2020scale,salzmann2020trajectron,li2021spatio,westny2023mtp,wang2023spatio} with \gls{GCN}~\cite{kipf2017semisupervised} and \gls{GAT}~\cite{velickovic2018graph} as some of the most commonly used architectures.

Incorporating map-based information in trajectory prediction methods, similar to inter-agent interactions, can be done in several ways.
Many works employ \glspl{CNN} to extract scene features, using rasterized maps~\cite{cui2019multimodal,messaoud2021trajectory}, or \gls{HD} semantic maps~\cite{zhao2019multi,park2020diverse,salzmann2020trajectron,li2021spatio,zhou2023query} as inputs.
With the availability of \gls{HD} maps increasing, there have been several proposals to make effective use of them, \eg, representing them as lane graphs~\cite{liang2020learning,gao2023dynamic,deo2022multimodal,zhou2023query}.
Lane graphs consist of a set of nodes representing the scene with additional semantic information, such as lane types, lane directions, and lane connectivity.
The benefit of using lane graphs is their natural inclusion into the \gls{GNN} framework---assuring that data modalities are consistently handled.
In this work, lane graphs are used as a means of incorporating map-based information into the model.

\subsection{Motion Constraints}
A potential issue with black-box models tasked with motion forecasting is that model outputs can be physically infeasible or lack the general characteristics of natural motion.
Recently, there have been several proposals for incorporating differential constraints within trajectory prediction models~\cite{cui2020deep,salzmann2020trajectron,li2021spatio,yue2022human,wen2022social,westny2023mtp,westny2023eval}, revealing two distinct approaches:
one in which the motion constraints are pre-defined~\cite{cui2020deep,salzmann2020trajectron,li2021spatio} and based on kinematic (or dynamic) single-track models~\cite{kong2015kinematic}\cite[p. 613]{lavalle2006planning}, the second being that the constraints are learned~\cite{yue2022human,wen2022social,westny2023mtp}, \eg, using neural ODEs~\cite{chen2018neuralode}.
The architecture in \cite{cui2020deep}, one of the first to explicitly incorporate kinematic constraints as part of a deep (\gls{CNN}-based) trajectory prediction model, illustrated the potential of the approach.
The effect of incorporating differential constraints was investigated in detail in~\cite{westny2023mtp}, revealing that their inclusion can significantly stabilize learning and improve prediction performance, most notably in terms of extrapolation.
This was continued in~\cite{westny2023eval}, where the authors evaluated various motion models in the context of learning-based trajectory prediction, concluding that simple dynamic models are preferable over more complex ones, and best combined with a 2\textsuperscript{nd} order numerical solver.
Importantly, none of the diffusion-based trajectory prediction methods mentioned previously~\cite{gu2022stochastic,chen2023equidiff,rempe2023trace} explicitly incorporate physical constraints as part of the learnable model.

In the computer-generated motion domain, generative networks are often regularized using geometric losses~\cite{shi2020motionet,petrovich2021action,tevet2023human} to encourage natural and coherent motion.
Of particular interest here is the diffusion model presented in~\cite{tevet2023human} as they modify the typical denoising objective~\cite{ho2020denoising} by not predicting the noise directly but instead the signal itself, meaning they can directly penalize the lack of physical characteristics in the model output.
Adopting a similar adjustment enables the inclusion of differential constraints directly in the model, offering the possibility of enforcing physically feasible trajectories.

%% file: sec/3_model.tex
\section{Trajectory Prediction Model}
\label{sec:model}
\begin{figure}[!t]
    \centering
    \includegraphics[width=\textwidth]{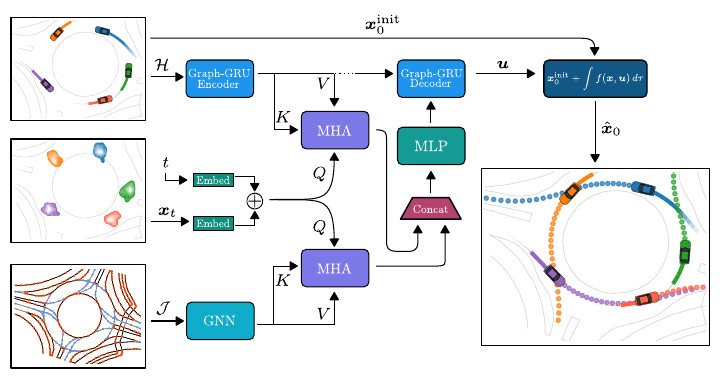}
    \caption{Schematic illustration of the proposed model. The model takes as input the condition $\cond = \{\history, \lanegraph\}$, the current latent state $\latent{\dstep}$, and diffusion step $\dstep$ to predict agent trajectories $\latent{0}$.
    The state history $\history$ up until the prediction time instant and the lane graph $\lanegraph$ are encoded using two \gls{GNN} modules.
    The diffusion step $\dstep$ is passed through a Fourier feature encoder~\cite{song2020score} and then summed with the embedded latent state.
    Next, the sum is passed through two \gls{MHA} mechanisms~\cite{vaswani2017attention}, one for each encoded condition part, and the resulting representations are concatenated and fused using an MLP to create a combined context encoding.
    The context encoding is then input into a Graph-GRU decoder (together with the last hidden state of the encoder) to compute the motion control inputs $\inp{}$.
    Using the computed control action and the agent states $\latent{0}^{\text{init}}$ at the prediction time instant, the predicted trajectories are solved for using numerical integration.
    }
    \label{fig:mdl}
\end{figure}

The proposed model, illustrated in \cref{fig:mdl}, consists of a \gls{GNN}-based diffusion model, adapted for multi-agent trajectory prediction. 
The model should synthesize future trajectories $\latent{0}^{1:\predhrz}$ of length $\predhrz$ for all agents $\agent\in\node$ currently in the scene at the prediction time instant given the condition $\cond$. 
The condition contains information about the agents' past trajectories, inter-agent interactions, and map-based information.
For brevity, $\latent{\dstep}$ is here on used to denote the full sequence at diffusion step $\dstep$.

\input{sec/3-2_gnn.tex}

\input{sec/3-1_ddpm_v2.tex}

\subsection{Differentially Constrained Predictions}
The differential constraints are enforced by having the learnable parts of the model $\model$ compute the 2-DOF control inputs $\inp{} = [u_x, u_y]$ to a motion model $f$ with continuous-time dynamics.
The output trajectory is then obtained by solving an initial value problem using Heun's 2\textsuperscript{nd} order method~\cite[p. 78]{ascher1998computer}.
For our purposes, we mainly employ two different motion models depending on the agent under investigation.
Wheeled vehicles (cars, trucks, motorcycles, \etc) are modeled as point masses, with longitudinal and lateral acceleration as inputs
\begin{align}
    \begin{split}
       & \dot{v}_x = u_x \\
       & \dot{v}_y = u_y \\
       & u_x^2 + u_y^2 \leq (\mu g)^2,
    \end{split}
\end{align}
where $\mu=0.7$ is the coefficient of road adhesion (slightly below dry asphalt)~\cite[p. 25]{wong2022theory}, and $g=9.81$ m/s\textsuperscript{2} is the acceleration due to gravity.
The inputs are elliptically constrained in a manner related to the tire friction ellipse~\cite[p. 45]{wong2022theory}, thereby accounting for vehicle capability constraints.
For pedestrians, which generally have more complex motion patterns, we apply first-order neural ODEs~\cite{chen2018neuralode}:
\begin{align}
    \begin{split}
       \dot{x} = f_x(\bm{x}, u_x) \\
       \dot{y} = f_y(\bm{x}, u_y),
    \end{split}
    \label{eq:pedestrian-ode}
\end{align}
where $\bm{x} = [x, y]$ is the state vector, and $f_x$ and $f_y$ are MLPs with learnable parameters where we decouple the state-transition equations to reduce the model complexity.
Notably, the inputs of \cref{eq:pedestrian-ode} are trained to be close to the ground truth velocities to improve the interpretation of their physical meaning.

%% file: sec/3-2_gnn.tex
\subsection{Enviroment-Aware Modeling}
\label{sec:mdl-eam}
Inter-agent relationships over $n$ time steps are modeled as sequences $\graph_1, \dots, \graph_n$ of graphs centered around a vehicle $\agent_0$ and each node represents a traffic participant.
Let $\graph_i = (\node_i, \edge_i)$ be a graph at time $i$, where $\node_i$ is the node set and $\edge_i$ is the edge set.
Additionally, Let $\feature{i} \in \R^{d_f}$ be the features of node $\agent \in \node_i$ and $\ew{\agent}{\neighbor}_i \in \R^{d_e}$ be the features of edge $(\agent, \neighbor) \in \edge_i$, where the features contain information about the agents' states, such as position and velocity.
It is important to note that the cardinality of the graphs within the observation window $\predhist$ may vary as a result of the arrival and departure of agents.
The exact observation history $\history$ of length $\predhist$ can be summarized as $\history = \{\graph_{1-h}, \dots \graph_{-1}, \graph_{0}\}$. %
We note that the exact graph structure is unknown during the decoding process, \ie, over the prediction horizon.
During this stage, the model utilizes the last observed graph $\graph_0$ as a proxy for the future graph structures.

In addition to inter-agent interactions, we draw inspiration from \cite{liang2020learning,deo2022multimodal} and model road-agent relationships using lane graphs, denoted by $\lanegraph = (\tilde{\node}, \tilde{\edge})$.
In contrast to the inter-agent graphs, the lane graphs are not temporal by nature but instead contain a spatial snapshot of the scene at the prediction time instant. 
The important distinction, however, is that the lane graphs additionally include \emph{environmental} nodes that hold information about road boundaries and lane lines. 
The lane graphs are constructed based on the geometric constraints of the current road topology and connect environmental nodes to agents through directed edges based on a nearest-neighbor criterion.
Together, the lane graph and the inter-agent graphs form the condition $\cond = \{\history, \lanegraph\}$.

\subsubsection{Graph Attention Network.}
\label{sec:mdl-gnn}
Both $\history$ and $\lanegraph$ are encoded using \glspl{GNN} based on the improved \gls{GAT} architecture (GATv2)~\cite{brody2022attentative}.
\gls{GAT} layers use an attention mechanism to compute a set of aggregation weights over the inclusive neighborhood $\ineigh{\agent} = \neigh{\agent} \cup \{\agent\}$ for each node $\agent$, enabling the \gls{GNN} to focus more on specific neighbors in the graph.
The attention coefficients are computed as
\begin{equation}
    \gattw{\tau} = \frac{
        \exp\left(\attn^\transpose{}
        \lrelu\left(
            \weightind{1} \hidden{}^\agent + \weightind{2} \hidden^\neighbor + \weightind{3} \ew{\agent}{\neighbor}
        \right)\right)
    }{
    \sum_{\upsilon \in \ineigh{\agent{}}}
        \exp\left(\attn^\transpose{}
        \lrelu\left(
            \weightind{1} \hidden{}^\agent + \weightind{2} \hidden^\upsilon + \weightind{3} \ew{\agent}{\upsilon}
        \right)\right)
    },
\end{equation}
where $\attn$ and $\weight_{(\cdot)}$ are learnable matrices.
The node update rule using the attention weights is then given by
\begin{equation}
    \updatedrep{} = \bm{b} + \gattw{\agent}\weightind{1} \hidden{}^\agent{} + \smashoperator{\sum_{\tau \in \neigh{\agent{}}}} \gattw{\tau} \weightind{2} \hidden{}^\tau,
\end{equation}
where $\hidden{}^\agent{}$ is the current embedding of node $\agent$, and $\bm{b}$ is a learnable bias term.
The employed \gls{GNN}, is a small extension of \gls{GAT}v2 where an additional weight matrix $\weightind{4}$ is introduced only for the center node in the update step:
\begin{equation}
    \updatedrep{} = \bm{b} + (\gattw{\agent}\weightind{1} + \weightind{4}) \hidden{}^\agent{} + \smashoperator{\sum_{\tau \in \neigh{\agent{}}}} \gattw{\tau} \weightind{2} \hidden{}^\tau,
\end{equation}
thereby introducing additional flexibility in how the representation of the center node is used.

\subsubsection{Graph-Gated Recurrent Unit.}
\label{sec:mdl-ggru}
To capture the temporal aspect of the data, \glspl{GGRU} are adopted.
The \gls{GGRU} is a recurrent unit that enables spatio-temporal interactions to be captured by replacing the linear mappings in the conventional \gls{GRU}~\cite{cho2014properties} with \gls{GNN} components~\cite{zhao2018deep,oskarsson2022temporal}.
The \glspl{GNN} take as input the representations for the specific node $\agent$ and the information of its neighborhood.
Intermediate representations are computed by two \glspl{GNN} as
\begin{subequations}
\label{eq:gnn_interm}
\al[gnn_interm_f]{
    \left[\grurepx{r} \| \grurepx{z} \| \grurepx{h} \right] &=
        \gnnf{i}\\
    \label{eq:gnn_interm_h}
    \left[\grureph{r} \| \grureph{z} \| \grureph{h} \right] &=
        \gnnh{i-1},
}
\end{subequations}
where $\|$ is the concatenation operation.
These are then used to compute the embedded state $\encrep{i}$ for time step $i$ as
\begin{subequations}
\begin{align}
	\label{eq:full_gru_update}
	\bm{r}_i^\agent &= \sigma(\grurepx{r} + \grureph{r} + \bm{b}_r) \\
	\bm{z}_i^\agent &= \sigma(\grurepx{z} + \grureph{z} + \bm{b}_z) \\
	\bm{n}_i^\agent &= \text{tanh}(\grurepx{n} + \bm{r}_i^\agent \odot \grureph{n} + \bm{b}_{n})\\
	\encrep{i} &= (\bm{1} - \bm{z}_i^\agent) \odot \bm{n}_i^\agent{} + \bm{z}_i^\agent \odot \encrep{i-1},
\end{align}
\end{subequations}
where the bias terms $\bm{b}_r$, $\bm{b}_z$, $\bm{b}_n$ are additional learnable parameters, $\odot$ the Hadamard product, and $\sigma$ is the sigmoid function. %
\glspl{GGRU} are used to encode the history $\history$ and to decode the motion control inputs $\inp{}$.

%% file: sec/3-1_ddpm_v2.tex
\subsection{Diffusion-Based Forecasting}
\label{sec:mdl-ddpm}
The underlying principle of diffusion models is to progressively perturb the observed data with a forward (diffusion) process, then recover the original data through a reverse process~\cite{sohl2015deep,ho2020denoising,song2021denoising,karras2022elucidating}.
These processes are commonly modeled through two Markov chains (see \cref{fig:graphical-model} for an example of how the process is modeled in this work).
To learn the reverse transition, a continuous-time forward transition from $\latent{0}$ to $\latent{\dstep}$ is first defined as in~\cite{jabri2022scalable}:
\begin{equation}
    \label{eq:forward-process}
    \latent{\dstep} = \sqrt{\schedule(\dstep)} \latent{0} + \sqrt{1 - \schedule(\dstep)} \noise,
\end{equation}
where $\noise\sim\normal(\zero, \eye)$, $\dstep\sim\uniform(0, 1)$ and $\schedule(\dstep)$ is a monotonically decreasing noise scheduling function.
The transition function given by \cref{eq:forward-process} allows for the direct sampling of $\latent{\dstep}$ at any arbitrary diffusion step $\dstep$ without the need for computation of the forward process step by step.
Selecting a sufficiently large number of diffusion steps $\dsteps$ and progressing in the diffusion direction ensures that $\latent{0}$ becomes fully corrupted, \ie, $\latent{1} \sim \normal(\zero, \eye)$.
However, in the context of multi-agent trajectory prediction, it is more effective to instead draw samples from a distribution that is specific to each agent.
To that end, we propose to adjust the forward process as 
\begin{equation}
    \label{eq:forward-process-mod}
    \latent{\dstep} = \sqrt{\schedule(\dstep)} \latent{0} + \left(1 - \sqrt{\schedule(\dstep)}\right) \latent{0}^{\text{init}} + \sqrt{1 - \schedule(\dstep)} \noise,
\end{equation}
where $\latent{0}^{\text{init}}$ refers to the agent states at the prediction time instant, the effect being that $\latent{1} \sim \normal(\latent{0}^{\text{init}}, \eye)$.
This method allows the network to tailor the reversal of the diffusion process according to the unique characteristics and states of each agent.

\subsubsection{Objective.}
Instead of predicting $\noise_\dstep$ as formulated in~\cite{ho2020denoising,song2021denoising}, the model learns to predict the clean signal, \ie, $\predlatent{0} = \model(\latent{\dstep}, \dstep, \cond)$ using the simple objective~\cite{ho2020denoising},
\begin{equation}
    \mathcal{L} = \min_\theta \mathbb{E}_{\latent{0} \sim \posterior(\latent{0}), \noise\sim\normal(\zero, \eye), \dstep\sim\uniform(0,1)} \lVert \latent{0} - \model(\latent{\dstep}, \dstep, \cond) \rVert^2_2
\end{equation}
This adaptation is inspired by the arguments made in~\cite{tevet2023human}, where the addition of geometric losses is used to enforce physical properties in a neural motion generation model.
Adopting this simple adjustment enables the inclusion of differential constraints by having the learnable parts of the model $\model$ compute the inputs to a motion model, a modification that has been an important part in several recent trajectory prediction models~\cite{cui2020deep,salzmann2020trajectron,li2021spatio,westny2023mtp,westny2023eval}. %

\subsubsection{Reverse Process.}
To generate samples from a learned model, we follow a series of (reverse) state transitions $\latent{1} \rightarrow \latent{1-\ddiff} \rightarrow \cdots \rightarrow \latent{0}$.
This is done by iteratively applying the denoising function $\model$ on each latent state $\latent{\dstep}$ and predicting the clean signal $\predlatent{0}$.
The current prediction is then diffused to $\latent{\dstep-\ddiff}$ for usage in the next iteration of the reverse process; a procedure which is repeated for $\dsteps$ steps until $\dstep=\varepsilon$, where $\varepsilon$ is a small positive number (typically, $\varepsilon\in O(10^{-3})$~\cite{karras2022elucidating}).
The sampling procedure can be conducted using various transition rules, \eg, DDPM~\cite{ho2020denoising}, DDIM~\cite{song2021denoising}, or EDM~\cite{karras2022elucidating}.
Importantly, we note that the diffusion steps are not necessarily equidistant in time, \ie, $\ddiff$ is not constant.
In practice, we follow \cite{karras2022elucidating} where the time steps are distributed according to
\begin{equation}
    \label{eq:time-step}
    t_{i < \dsteps} = \left(\schedule_{\rm max}^\frac{1}{\rho} + \frac{i}{\dsteps - 1}(\schedule_{\rm min}^\frac{1}{\rho} - \schedule_{\rm max}^\frac{1}{\rho}) \right)^\rho, \quad t_\dsteps = 0,
\end{equation}
and set $\rho=3$, meaning we are taking smaller steps as $i$ approaches $\dsteps$.
This modification allows for more fine-grained control of the reversal process near the end of the diffusion process.

\subsubsection{Interaction-Aware Guidance.}
\label{sec:mdl-iag}
An important attribute of trajectory prediction models is their interaction-aware capabilities.
In this work, such interactions are modeled through the use of graphs, where node connectivity is described by the edge set $\edge$.
Removing elements from the edge set is a simple way to disable message-passing between nodes and offers an interesting approach to investigating the effect of inter-agent interactions on the prediction outcome during test time.
Let $\edge' = \{(\agent, \agent) \mid \agent \in \node\}$ denote the set of all self-loop edges in the graph $\graph$ and $\cond' = (\history', \lanegraph)$ the corresponding condition where each element in $\history'$ has $\edge'$ as its edge set.
Inspired by \emph{classifier-free guidance}~\cite{ho2021classifierfree}, we seek to investigate how node connectivity can be used to guide the denoising process by modifying the prediction rule according to
\begin{equation}
    \label{eq:guidance-signal}
    \predlatent{0} = (1 - w) \cdot \model(\latent{\dstep}, \dstep, \cond') + w \cdot \model(\latent{\dstep}, \dstep, \cond),
\end{equation}
where $w \in [0, 1]$ is a weighting factor that controls the influence of inter-agent interactions on the predicted trajectories.
To prevent the model from becoming overly dependent on the guidance signal, edges are randomly removed during training, thus enabling a transition to a guidance-free model during inference.

%% file: sec/4_results.tex
\section{Evaluations \& Results}
\label{sec:results}
The proposed model is evaluated on two real-world datasets and assessed both quantitatively and qualitatively.
The metrics used for the evaluation are:
\begin{enumerate}
    \item \gls{ADE}: The average Euclidean distance in meters between the predicted and ground truth trajectories. %
    \item \gls{FDE}: The Euclidean distance in meters between the predicted and ground truth trajectories at the final time step.
    \item \gls{MR}: The percentage of predictions where the predicted trajectory at the final time step is not within $2.0$~m from the ground truth.
\end{enumerate}
Since our model has no explicit modes, we take the mean of 6 samples to calculate the metrics in order to make a fair comparison with the baselines.
Apart from the proposed model, we evaluate two additional variants:
\begin{enumerate}[label=(\alph*)]
    \item The proposed model but without the motion model $f$ (denoted `'$\setminus f$`' in tables), \ie, the decoder computes the predicted trajectory directly. %
    \item Like (a) with a closed-loop refinement step (denoted `'$\circlearrowleft$`' in tables). Using the output of the diffusion model as a reference trajectory, we apply a simple closed-loop control scheme based on the pure-pursuit algorithm~\cite{coulter1992implementation} with a speed-dependent lookahead horizon to refine the predicted trajectory (see \cref{app:ppc} for more details).
\end{enumerate}

\subsubsection{Datasets.}
Experiments are conducted using the \highd{}~\cite{highDdataset} and \round{}~\cite{rounDdataset} datasets.
The datasets contain trajectories from various locations in Germany captured at a frequency of $25$~Hz.
The data is downsampled by a factor of $5$, resulting in a sampling time of $0.2$~s.
The maximum length of the observation window is set to $3$~s ($h=15$), while the prediction horizon is $5$~s ($N=25$).
The preprocessed \highd{} and \round{} datasets consist of 100~404 and 29~248 samples, respectively.
We allocate 80\% of the total samples for training, 10\% for validation, and 10\% for testing.
Contrary to the \round{} dataset, the \highd{} dataset does not contain any explicit lane-graph information.
In those cases, the model only uses the trajectory information.
This also provides an opportunity to investigate the practical significance of the map encoder.

\subsubsection{Implementation Details.} 
The model was implemented using the PyTorch library~\cite{paszke2019pytorch} and trained using the \emph{AdamW} optimizer~\cite{loshchilov2018decoupled} with a learning rate of $0.0005$ and a batch size of $512$ on a single NVIDIA GeForce RTX 3090 GPU.
The learning rate was decayed using a cosine annealing learning rate scheduler~\cite{loshchilov2017sgdr}.
The diffusion process is modeled using a linear scheduling function $\schedule(\dstep) = 1 - \dstep$.
Sample generation was conducted using the 1\textsuperscript{st} order method from \cite{karras2022elucidating}, which we found worked best in practice.
Importantly, we only require $\dsteps=2$ diffusion steps to obtain highly competitive samples which we use for the evaluation.

\input{tables/drone.tex}

\subsection{Quantitative Results}
\label{sec:quantitative-results}
The quantitative results are presented in \cref{tab:drone-res}.
Performance is reported for different values of the weighting factor $w$ from \cref{eq:guidance-signal}, used to highlight the importance of including inter-agent interactions.

The proposed model outperforms several established methods on both datasets, achieving the lowest \gls{ADE}, \gls{FDE}, and \gls{MR} scores.
Interestingly, the model performs well on the \highd{} dataset, despite the lack of map information.
Compared to the performance reported in \cite{chen2023equidiff} that struggled with making accurate predictions on long-term trajectories using a diffusion-based model, our method does not share the same limitations.
This is likely a result of the incorporation of differential constraints as this helps guide the model to generate more realistic and high-quality samples.
This is further corroborated by the results of the model that excludes the motion model $f$, which performs much worse than its counterpart.

Inspired by the diffusion model in \cite{rempe2023trace}, we also include a variant of our model that uses a closed-loop refinement step.
The results show that the refinement step has a slight positive effect on the prediction accuracy compared to the model without $f$, which is more pronounced on the highway dataset, but still not enough to outperform the proposed architecture. 
A notable drawback of this approach is the added complexity of adding a closed-loop control scheme, which requires additional tuning and testing depending on the specific application.

It is worth noting that the model outperforms \mtp{}~\cite{westny2023mtp} on both datasets, which is a strong baseline for trajectory prediction and shares many architectural similarities with our model.
We hypothesize that this could be owing to our choice of motion model, which has a more physically grounded objective.

 \begin{figure}[!t]
    \centering
    \includegraphics[width=\textwidth]{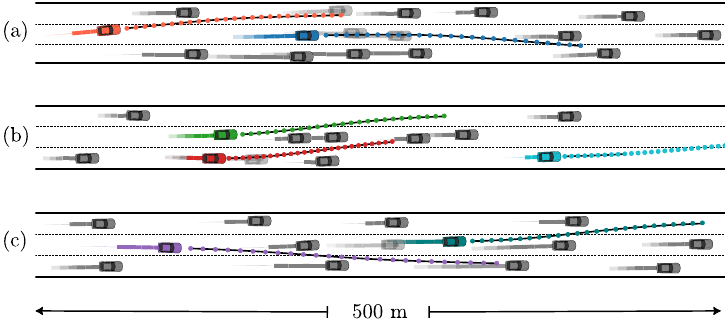}
    \caption{Three example predictions on the \highd{}~\cite{highDdataset} test set. The vehicles in the plots are used to represent the agents in the scene at the prediction time instant. The predicted trajectories are shown with the same colored scatter plots as the agent under investigation, while the ground truth is shown with a solid black line. For visual clarity, we only show the predicted trajectories for vehicles performing a lane change (the model predicts lane-keeping maneuvers with equal accuracy).
    }
    \label{fig:hwc}
\end{figure}
\begin{figure}[!t]
    \centering
    \includegraphics[width=\textwidth]{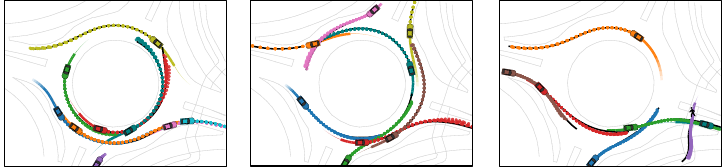}
    \caption{Three example predictions on the \round{}~\cite{rounDdataset} test set. The predicted trajectories are shown with the same colored scatter plots as the agent under investigation, while the ground truth is shown with a solid black line.
    What is interesting to note is the accurate prediction of the pedestrian in the rightmost plot.
    Since the state-transition dynamics are parametrized using neural ODEs, it illustrates that the model has learned a motion model that is representative of the real-world data.
    }
    \label{fig:rdc}
\end{figure}

\subsection{Qualitative Assessment}
Given the encouraging performance shown in \cref{sec:quantitative-results}, it is interesting to investigate how predictions look visually.
In \cref{fig:hwc}, we present example predictions using the highway data.
Even though the average predicted distance is 200--300~m ahead of the observed time instant, the predictions are highly accurate and closely in line with the ground truth trajectories.
Interestingly, although the dataset contains a large number of vehicles performing lane-keeping maneuvers, the model predicts lane changes with equal accuracy.
In \cref{fig:rdc}, we present representative predictions using the roundabout data.
The model is mostly accurate in predicting future trajectories, even when various agents are involved.

\subsubsection{Multimodality.}
\begin{figure}[!t]
    \centering
    \includegraphics[width=\textwidth]{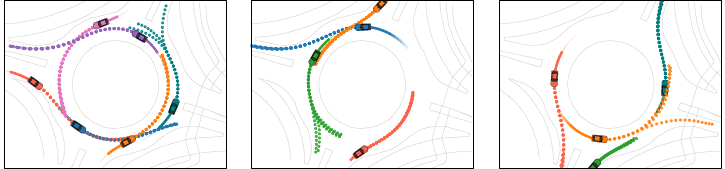}
    \caption{Three example predictions on the \round{}~\cite{rounDdataset} test set. 
    The multiple predicted trajectories for a specific agent are shown with same-colored scatter plots.
    The scenarios were chosen to illustrate the model's ability to predict a diverse set of future trajectories. %
    Each prediction is based on a unique sample from the diffusion process, thereby showcasing that the model has implicitly learned to capture the multimodal nature of the data.
    }
    \label{fig:mm}
\end{figure}

In \cref{fig:mm}, three example predictions using roundabout data are presented---showcasing the model's ability to provide multimodal predictions.
Although the model has no explicit representation of modes, it is interesting to see that the model can still generate multiple candidate trajectories given the same condition.
We note that to generate a diverse set of predictions, the model needs to be sampled multiple times following the reverse process outlined in \cref{sec:mdl-ddpm}, which is a limitation that might need to be addressed in future work.
However, our model still only requires very few diffusion steps to generate high-quality samples.
It should also be noted that the multimodal nature of the predictions was found to be more pronounced in the roundabout scenarios.
This is likely due to the increased complexity of the scenarios and the larger number of possible future trajectories given the same condition.

\subsubsection{Interaction-Aware Guidance.}
\begin{figure}[!t]
    \centering
    \includegraphics[width=\textwidth]{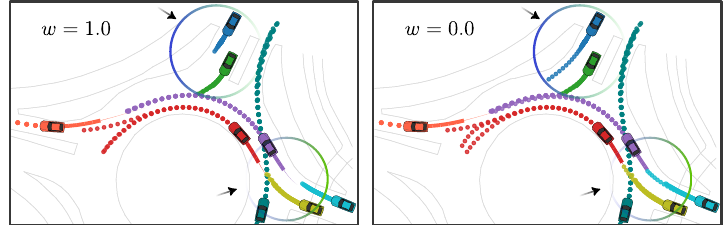}
    \caption{Example predictions on the \round{}~\cite{highDdataset} dataset for different values of interaction-aware guidance signal strength $w$, where $w=0.0$ removes all node-to-node message passing, while $w=1.0$ preserves the original graph structure. 
    As was discussed in \cref{sec:quantitative-results}, letting $w<1.0$ has limited practical use when assessing overall prediction accuracy.
    However, as the circles highlight in the figure, removing inter-agent interactions gives rise to more \emph{egocentric} predictions: agents that typically slow down to yield to other agents in the scene, now enter into the roundabout without any regard for other agents.
    Vehicles that are already in the roundabout are seemingly unaffected by this modification which is supported by the traffic rules, something the model has learned.
    }
    \label{fig:ia-guide}
\end{figure}

Originally, guidance-based sampling\cite{ho2021classifierfree} was proposed to enable diffusion models that could generate both unconditional and conditional samples based on the availability of a guidance signal (typically image labels).
In this work, we investigate how node connectivity can guide the prediction process.
When $w=0.0$, the model only uses graphs with self-loops, effectively computing predictions based on the individual agent's trajectory.
When $w=1.0$, the model instead considers the original graph structure.
The test results show that using $w<1.0$ worsens prediction performance, arguably due to its usefulness in modeling inter-agent interactions.
Still, we found that removing edges from the graphs during \emph{training} improved validation accuracy, a generalization effect that has been observed in other graph-based models~\cite{rong2020drop}.
However, when observing the trajectories for $w<1.0$, we found; perhaps unsurprisingly, that the model generates predictions that more so reflect a behavior that is more centered around the individual goal of the agent in question, as is illustrated in \cref{fig:ia-guide}.
We hypothesize that this could be used for predicting the behavior of less cooperative agents or drivers with a more aggressive driving style---potentially of interest in the context of robust motion planning.

%% file: tables/drone.tex
\begin{table}[!t]
    \centering
    \caption{Quantitative performance on the \highd{}~\cite{highDdataset}~\textbf{\scriptsize(a)} and \round{}~\cite{rounDdataset}~\textbf{\scriptsize(b)} datasets, based on the metrics in \cite{westny2023mtp}. The best results are in bold, and the second best are underlined.}
    \label{tab:drone-res}
    \begin{minipage}{0.5\textwidth}
        \centering
        \subcaption{Highway}
        \label{tab:highd-res}
        \resizebox{\columnwidth}{!}{%
        {
            \setlength{\tabcolsep}{6pt} %

            \begin{tabular}{l c c c}
                \toprule
                Method & ADE $\downarrow$& FDE $\downarrow$& MR $\downarrow$\\
                \midrule
                Constant Acc. & $0.78$ & $2.63$ & $0.55$ \\ 
                S-LSTM~\cite{alahi2016social} & $0.41$ & $1.49$ & $0.22$ \\ 
                CS-LSTM~\cite{deo2018convolutional} & $0.39$ & $1.38$ & $0.19$ \\
                \gplusplus~\cite{li2019grip} & $0.38$ & $1.49$ & $0.18$ \\
                \mmtf{}~\cite{liu2021multimodal} & $0.39$ & $1.13$ & $0.15$ \\
                \tplusplus{}~\cite{salzmann2020trajectron} & $0.44$ & $1.62$ & $0.23$ \\ 
                \mtp~\cite{westny2023mtp} & $ 0.30 $ & $ 1.07 $ & $0.13$ \\
                \midrule
                Ours ($w=0.0$) & $0.39$ & $1.43$ & $0.23$ \\
                Ours ($w=0.4$) & $\underline{0.29}$ & $1.16$ & $0.15$ \\
                Ours ($w=0.8$) & $\underline{0.29}$ & $\underline{1.01}$ & $\underline{0.12}$ \\
                Ours ($w=1.0$) & $\bm{0.28}$ & $\bm{0.99}$ & $\bm{0.11}$ \\ %
                \midrule
                Ours$\setminus f$ ($w=1.0$) & $0.46$ & $1.27$ & $0.16$ \\
                Ours$\circlearrowleft$ ($w=1.0$) & $0.36$ & $1.25$ & $0.17$ \\
                \bottomrule
            \end{tabular}
        }}
    \end{minipage}%
    \begin{minipage}{0.5\textwidth}
        \centering
        \subcaption{Roundabout}
        \label{tab:rounD-res}

        \resizebox{\columnwidth}{!}{%
        {
            \setlength{\tabcolsep}{6pt} %

            \begin{tabular}{l c c c}
                \toprule
                Method & ADE $\downarrow$ & FDE $\downarrow$ & MR $\downarrow$ \\
                \midrule
                Constant Acc. & $4.83$ & $16.2$ & $0.95$ \\
                S-LSTM~\cite{alahi2016social} & $1.20$ & $3.47$ & $0.56$ \\
                CS-LSTM~\cite{deo2018convolutional} & $1.19$ & $3.57$ & $0.60$ \\
                \gplusplus~\cite{li2019grip} & $1.11$ & $3.19$ & $0.52$ \\
                \mmtf{}~\cite{liu2021multimodal} & $1.29$ & $3.50$ & $0.59$ \\
                \tplusplus{}~\cite{salzmann2020trajectron} & $1.09$ & $3.53$ & $0.54$ \\
                \mtp~\cite{westny2023mtp} & $0.96$ & $2.95$ & $\bm{0.46}$\\
                \midrule
                Ours ($w=0.0$) & $1.23$ & $3.99$ & $0.64$ \\
                Ours ($w=0.4$) & $1.01$ & $3.31$ & $0.57$ \\
                Ours ($w=0.8$) & $\underline{0.86}$ & $\underline{2.82}$ & $\underline{0.49}$ \\
                Ours ($w=1.0$) & $\bm{0.84}$ & $\bm{2.73}$ & $\bm{0.46}$ \\ 
                \midrule
                Ours$\setminus f$ ($w=1.0$) & $0.96$ & $3.10$ & $0.52$ \\
                Ours$\circlearrowleft$ ($w=1.0$) & $0.96$ & $2.99$ & $0.49$ \\
                \bottomrule
            \end{tabular}%

        }
        }
    \end{minipage}
\end{table}

%% file: sec/A_experiments.tex
\section{Extended Implementation Details}
As noted in the main paper, we evaluated the proposed model on two large-scale real-world datasets: \highd{}~\cite{highDdataset} and \round{}~\cite{rounDdataset}.
The proposed method and its variants were implemented using the PyTorch library~\cite{paszke2019pytorch} and trained using the \emph{AdamW} optimizer~\cite{loshchilov2018decoupled} with a learning rate of $0.0005$ and a batch size of $512$ on a single NVIDIA GeForce RTX 3090 GPU.
The learning rate was decayed using a cosine annealing learning rate scheduler~\cite{loshchilov2017sgdr}.
Graph neural network layers (GATv2~\cite{brody2022attentative}) were implemented using the \emph{PyTorch Geometric} library~\cite{pyg2019}.
Detailed parameter configurations for the model are provided in \cref{tab:params}.
The (diffusion) time embedding module consists of a Fourier feature encoder~\cite{song2020score} followed by a 2-layered MLP and the latent state embedding consists of a single linear layer.
The context fusion encoder consists of a 2-layered MLP.
The hidden dimension of all MLPs (linear layers) is the same as the encoder/decoder hidden dimension.
Whenever applicable, we used the SiLU (Swish) activation function~\cite{hendrycks2016gaussian}. %

\input{tables/params.tex}

\input{tables/feats.tex}

\subsection{Methods Compared}
The following models were included in the comparative study:
\begin{itemize}[label={\scriptsize\raisebox{0.1ex}{\ding{69}}}]
	\item Constant Acc.: Open-loop model assuming constant acceleration.
	\item S-LSTM~\cite{alahi2016social}: Uses an encoder--decoder network  based on LSTM for trajectory prediction.
	Interactions are encoded using social pooling tensors.
	\item CS-LSTM~\cite{deo2018convolutional}:
	Similar to S-LSTM, but instead applies \gls{CNN} to learn interactions from the pooling tensors.
	\item \gplusplus~\cite{li2019grip}: Encodes interactions using a graph network and generates trajectories with an RNN-based encoder--decoder.
	\item \mmtf{}~\cite{liu2021multimodal}: Transformer-based model for multimodal trajectory prediction.
	Interactions are encoded using multiple stacked Transformers. Generates several candidate proposals and selects the most likely one using a learned scoring function.
	\item \tplusplus{}~\cite{salzmann2020trajectron}: \gls{GNN}-based recurrent model.
	Performs trajectory prediction by a CVAE-based model together with hard-coded kinematic constraints.
	\item \mtp~\cite{westny2023mtp}: \gls{GNN}-based recurrent model. Computes trajectories with an MDN-based decoder and learned differential motion constraints.
\end{itemize}
All results were obtained from the extended study and experiments conducted in \cite{westny2023mtp} where all methods were modified to make use of the same input features (see \cref{tab:node_feat}), including edge weights for graph-based methods.
To make a fair comparison, our proposed method did not use the roundabout-specific features but instead the lane graph, with \emph{lane specific} features as node features.

\subsection{Modeling Practicalities}
Recall that wheeled vehicles (cars, trucks, motorcycles, \etc) in this work are modeled as point masses, with longitudinal and lateral acceleration as inputs
\begin{align}
    \begin{split}
       & \dot{v}_x = u_x \\
       & \dot{v}_y = u_y \\
       & u_x^2 + u_y^2 \leq (\mu g)^2.
    \end{split}
\end{align}
In practice, computed inputs are updated based on a conditional statement that utilizes the fact that input vector $\inp{}$ can be expressed in a polar coordinate frame
\begin{subequations}
\begin{align}
	\rho &= \sqrt{u_x^2 + u_y^2} \\
	\phi &= \arctan\left(\frac{u_y}{u_x}\right).
\end{align}
\end{subequations}
Since the upper bound on the input signals only affects the magnitude $\rho$ of the input vector, the constraint can be enforced by updating it according to
\begin{equation}
	\rho' \coloneqq
	\begin{cases}
		\rho \quad &\rho \leq \mu g \\
		\mu g \quad &\text{otherwise} 
	\end{cases}
\end{equation}
which in practice is achieved by the use of a HardTanh activation function with an upper bound of $\mu g$.
The (possibly) scaled magnitude is then used to recompute the input signals to the simulation model as 
\begin{subequations}
	\begin{align}
		u_x &= \rho' \cos(\phi) \\
		u_y &= \rho' \sin(\phi),
	\end{align}
\end{subequations}
thus retaining the direction of the original input vector.

\subsection{Pure-Pursuit Controller}
\label{app:ppc}
One proposed variant of our method incorporates a closed-loop refinement step where the output of the diffusion model is used as a reference trajectory.
The closed-loop control scheme is based on the pure-pursuit algorithm~\cite{coulter1992implementation} and includes a speed-dependent lookahead horizon.
The method has a long history of success in the robotics community, much attributed to its simplicity and satisfactory performance~\cite{paden2016survey}.
The algorithm is based on fitting a semi-circle from the vehicle's current configuration to a (pursuit) point on the reference path ahead of the vehicle by the lookahead distance $\lookahead$.
The curvature $\curvature$ of the circle is given by
\begin{equation}
	\curvature = \frac{2 p}{\lookahead}, \quad 
	p = \begin{bmatrix}
        x_{\text{ref}} - x \\
        y_{\text{ref}} - y
    \end{bmatrix}^T
    \begin{bmatrix}
        \sin(\psi) \\
        \cos(\psi)
    \end{bmatrix}
\end{equation}
where $\psi$ is the yaw angle.
In order to adapt the control law to different scenarios, a variable lookahead that depends on the vehicle's current speed $v$ was used:
\begin{equation}
    \lookahead_v = \lookahead + \lookaheadconstant \cdot |v|,
    \label{eq:lookahead}
\end{equation}
where $\lookaheadconstant$ is a constant that determines the rate of increase of the lookahead distance with speed $v$.
Using the computed control action, the vehicles' states are simulated forward using a curvature-input unicycle model:
\begin{align}
    \begin{split}
    \dot{x} &= \hat{v} \cos(\psi),\\
    \dot{y} &= \hat{v} \sin(\psi),\\
    \dot{\psi} &= \hat{v} \curvature,
    \end{split}
\end{align}
where $\hat{v}$ is the vehicle's predicted speed (taken as the norm of the velocities in $\predlatent{0}$).
The values of $\lookahead$ and $\lookaheadconstant$ were chosen based on a grid search over the validation set, taken as $1.5$ and $0.5$, respectively.

%% file: tables/params.tex
\begin{table}[!h]
    \centering
    \caption{Model hyperparameters.}
    \label{tab:params}
    \vspace{-0.1in}
    \begin{minipage}{0.5\textwidth}
        \centering
        \subcaption{Highway}
        \label{tab:highd-params}
        \resizebox{\columnwidth}{!}{%
        {
            \setlength{\tabcolsep}{6pt} %

            \begin{tabular}{l c c c c c}
                \toprule
                Specification & Encoder & Decoder & Lane & MHA & $f_{\rm ped}$ \\
                \midrule
                Dim. & $512$ & $512$ & --- & $512$ & --- \\
                Layers & $1$ & $1$ & --- & --- & --- \\
                Heads & $1$ & $1$ & --- & $4$ & --- \\
                \bottomrule
            \end{tabular}%
        }
        }
    \end{minipage}%
    \begin{minipage}{0.5\textwidth}
        \centering
        \subcaption{Roundabout}
        \label{tab:rounD-params}

        \resizebox{\columnwidth}{!}{%
        {
            \setlength{\tabcolsep}{6pt} %

            \begin{tabular}{l c c c c c}
                \toprule
                Specification & Encoder & Decoder & Lane & MHA & $f_{\rm ped}$ \\
                \midrule
                Dim. & $128$ & $128$ & $128$ & $128$ & $32$ \\
                Layers & $1$ & $1$ & $3$ & --- & $2$ \\
                Heads & $1$ & $1$ & $4$ & $4$ & --- \\
                \bottomrule
            \end{tabular}%

        }
        }
    \end{minipage}
\end{table}

%% file: tables/feats.tex
\begin{table}[!h]
	\caption{Node features used in the experiments. Edge weights were also included in the form of Euclidean distance between connected nodes.}
	\label{tab:node_feat}
	\centering
    \resizebox{0.7\columnwidth}{!}{%
        {
    \setlength{\tabcolsep}{10pt}
	\begin{tabular}{c l l}
		\toprule
		Feature & Description & Unit\\
		\midrule
		$x$ & Longitudinal coordinate & m\\
		$y$ & Lateral coordinate & m\\
		$v_x$ & Instantaneous longitudinal velocity & m/s \\
		$v_y$ & Instantaneous lateral velocity & m/s\\
		$a_x$ & Instantaneous longitudinal acceleration & m/$\text{s}^2$ \\
		$a_y$ & Instantaneous lateral acceleration & m/$\text{s}^2$\\
		$\psi$ & Yaw angle  & rad \\
		\midrule
		\multicolumn{3}{c}{\emph{Highway specific} \cite{westny2021vehicle}}\\
		$d_y$ & Lateral deviation from the current lane centerline & $[-1, 1]$\\
		$d_r$ & Lateral deviation from the road center & $[-1, 1]$\\
		\midrule
		\multicolumn{3}{c}{\emph{Roundabout specific} \cite{westny2023mtp}}\\
		$r$ & Euclidean distance from the roundabout center & m\\
		$\theta$ & Angle relative to the roundabout center & rad\\
        \midrule
		\multicolumn{3}{c}{\emph{Lane specific}}\\
		$x_{\lanegraph}$ & Longitudinal coordinate & m\\
		$y_{\lanegraph}$ & Lateral coordinate & m\\
		\bottomrule
	\end{tabular}
        }
    }
\end{table}

%% file: sec/B_diffusion.tex
\section{Diffusion Model Transition Rules}
The formulation of diffusion models can be done differently depending on how the forward and reverse processes are modeled: Either through a Markovian chain~\cite{ho2020denoising}, a non-Markovian process~\cite{song2021denoising}, a stochastic differential equation (SDE)~\cite{song2020score}, or as a (deterministic) probability flow ODE~\cite{song2020score}.
Regardless of the formulation, all models can be trained using the same forward transition from $\latent{0}$ to $\latent{\dstep}$:
\begin{equation}
    \latent{\dstep} = \sqrt{\schedule(\dstep)} \latent{0} + \textcolor{eccvblue}{\left(1 - \sqrt{\schedule(\dstep)}\right) \latent{0}^{\text{init}}} + \sqrt{1 - \schedule(\dstep)} \noise,
\end{equation}
where $\noise\sim\normal(\zero, \eye)$, $\dstep\sim\uniform(0, 1)$ and $\schedule(\dstep)$ is a monotonically decreasing noise scheduling function.
As noted in the main text, we follow a series of (reverse) state transitions $\latent{1} \rightarrow \latent{1-\ddiff} \rightarrow \cdots \rightarrow \latent{0}$ to generate samples from the learned model.
This is done by iteratively applying the denoising function $\model$ on each latent state $\latent{\dstep}$ and predicting the clean signal $\predlatent{0}$.
The sampling procedure can be conducted using various transition rules, \eg, DDPM~\cite{ho2020denoising}, DDIM~\cite{song2021denoising}, or EDM~\cite{karras2022elucidating}.
For completeness, a brief overview of these transition rules is provided.
In order to express all transition rules according to their original proposal using the same variables, we first define the following quantities:
\begin{align}
     \alphabart{\dstep} &\coloneqq \schedule(\dstep) \\
     \alphat{\dstep} &\coloneqq \frac{\alphabart{\dstep}}{\alphabart{\dstep-\ddiff}} \\
     \betat{\dstep} &\coloneqq 1 - \alphat{\dstep}.
\end{align}

\subsection{DDPM}
Diffusing $\predlatent{0}$ to retrieve $\latent{\dstep-\ddiff}$ is achieved by sampling from the posterior distribution when conditioned on $\latent{0}$,
\begin{equation}
    \posterior(\latent{\dstep-\ddiff} | \latent{\dstep}, \latent{0}) = \normal(\latent{\dstep-\ddiff}; \posteriormu(\latent{\dstep}, \latent{0}), \posteriorvar \eye),
\end{equation}
where the mean and variance of the posterior is given by 
\begin{equation}
    \posteriormu(\latent{\dstep}, \latent{0}) = \frac{\sqrt{\alphabart{\dstep-\ddiff}}\betat{\dstep}}{1 - \alphabart{\dstep}}\latent{0} + \frac{\sqrt{\alphat{\dstep}}(1 -\alphabart{\dstep-\ddiff})}{1 - \alphabart{\dstep}}\latent{\dstep}, \qquad 
    \posteriorvar = \frac{1 - \alphabart{\dstep-\ddiff}}{1 - \alphabart{\dstep}} \betat{\dstep}.
\end{equation}
A notable limitation of the DDPM approach is the prolonged sampling time, attributed to the number of reversal steps $\dsteps$ needed to produce high-quality samples.
Additionally, given its Markovian nature, time steps are assumed equidistant, which can lead to a suboptimal reversal process.

\subsection{DDIM}
To address the limitations of the DDPM, the DDIM model was proposed~\cite{song2021denoising}.
The DDIM model introduces a non-Markovian process, allowing for more flexible time steps and an accelerated sampling procedure,
\begin{equation}
    \posterior(\latent{\dstep-\ddiff} | \latent{\dstep}, \latent{0}) = \normal(\latent{\dstep-\ddiff}; \varposteriormu(\latent{\dstep}, \latent{0}), \varposteriorvar \eye),
\end{equation}
where
\begin{align}
    \begin{split}
    \varposteriormu(\latent{\dstep}, \latent{0}) &= \sqrt{\alphabart{\dstep-\ddiff}}\latent{0} + \sqrt{1 - \alphabart{\dstep-\ddiff} - \varposteriorvar} \cdot \frac{\latent{\dstep} - \sqrt{\alphabart{\dstep}}\latent{0}}{\sqrt{1 - \alphabart{\dstep}}} \\
    \varposteriorvar &= \eta^2 \frac{1 - \alphabart{\dstep-\ddiff}}{1 - \alphabart{\dstep}} \left(1 -\frac{\alphabart{\dstep}}{\alphabart{\dstep-\ddiff}}\right), \qquad \eta \in [0,1]
    \end{split}
\end{align}
where $\eta$ is a constant used to control the stochasticity of the reverse process (setting it to $1$ recovers the DDPM).

\subsection{EDM}
The EDM~\cite{karras2022elucidating} sampling procedure relies on the SDEs of Song \etal~\cite{song2020score}:
\begin{align}
    \begin{split}
    \text{d}\latent{\pm} &= \underbrace{-\dot{\schedule}(\dstep)\schedule(\dstep) \nabla_{\latent{}}\log\joint(\latent{} ; \schedule(\dstep)) \ \text{d}t}_{\text{probability flow ODE}} \\
    &\pm \underbrace{\lambda(\dstep)\schedule(\dstep)^2 \nabla_{\latent{}}\log\joint(\latent{} ; \schedule(\dstep)) \ \text{d}t + \sqrt{2 \lambda(\dstep)} \schedule(\dstep) \ \text{d}w_{\dstep}}_{\text{Langevin diffusion SDE}}
\end{split}
\end{align}
where $w_{\dstep}$ is the standard Wiener process and $\lambda(\dstep)$ expresses the rate at which existing noise is replaced with new noise~\cite{karras2022elucidating}.
The $\pm$ sign in the equation illustrates which SDE to solve for the forward or reverse process.
In \cite{karras2022elucidating}, a stochastic 2\textsuperscript{nd} order solver was proposed, shown to be more efficient than the transition rules used in DDPM and DDIM for image generation benchmarks.
Although the EDM sampler was used in this work to generate trajectories from the learned model, the 2\textsuperscript{nd} order correction step was omitted, a simplification that we found to be sufficient for our purposes.

Experimenting with the three sampling methods outlined above, it was found that using the EDM sampler gave the best prediction performance.
Still, the DDIM sampler was not far behind (for the same number of reversal steps).